\documentclass[letterpaper, 10pt, conference]{ieeeconf}
\IEEEoverridecommandlockouts
\makeatother

\usepackage{cite}
\usepackage{amsmath,amssymb,amsfonts}
\usepackage{algorithmic}
\usepackage{graphicx}
\usepackage{textcomp}
\usepackage{amssymb}
\usepackage{multirow}
\usepackage[table]{xcolor}
\makeatletter
\let\NAT@parse\undefined
\makeatother
\usepackage[colorlinks,
            linkcolor=red,
            anchorcolor=blue,
            citecolor=green]{hyperref}
\def\BibTeX{{\rm B\kern-.05em{\sc i\kern-.025em b}\kern-.08em
    T\kern-.1667em\lower.7ex\hbox{E}\kern-.125emX}}
\begin{document}

\title{\LARGE \bf MMFusion: A Generalized Multi-Modal Fusion Detection Framework*}
\author{Leichao Cui$^{1}$, Xiuxian Li$^{1}$, Min Meng$^{1}$, and Xiaoyu Mo$^{2}$
\thanks{*This work was supported by the National Natural Science Foundation of China under Grant 62003243 and Grant 62103305, and the Shanghai Municipal Science and Technology Major Project, No. 2021SHZDZX0100.}
\thanks{$^{1}$Department of Control Science and Engineering, College of Electronics and Information Engineering, and the Shanghai Research Institute for Intelligent Autonomous Systems, Tongji University, Shanghai 201800, China {\tt\small 2130715@tongji.edu.cn, 
        xxli@ieee.org,
        mengmin@tongji.edu.cn}}%
\thanks{$^{2}$School of Mechanical and Aerospace Engineering, Nanyang Technological University, 639798, Singapore.
        {\tt\small xiaoyu006@e.ntu.edu.sg}}%
}

\maketitle

\begin{abstract}
LiDAR point clouds have become the most common data source in autonomous driving. However, due to the sparsity of point clouds, accurate and reliable detection cannot be achieved in specific scenarios. Because of their complementarity with point clouds, images are getting increasing attention. Although with some success, existing fusion methods either perform hard fusion or do not fuse in a direct manner. In this paper, we propose a generic 3D detection framework called MMFusion, using multi-modal features. The framework aims to achieve accurate fusion between LiDAR and images to improve 3D detection in complex scenes. Our framework consists of two separate streams: the LiDAR stream and the camera stream, which can be compatible with any single-modal feature extraction network. The Voxel Local Perception Module in the LiDAR stream enhances local feature representation, and then the Multi-modal Feature Fusion Module selectively combines feature output from different streams to achieve better fusion. Extensive experiments have shown that our framework not only outperforms existing benchmarks but also improves their detection, especially for detecting cyclists and pedestrians on KITTI benchmarks, with strong robustness and generalization capabilities. Hopefully, our work will stimulate more research into multi-modal fusion for autonomous driving tasks. 
\end{abstract}

\section{Introduction}

 \begin{figure}[thpb]
      \centering
      
      \includegraphics[width=0.43\textwidth]{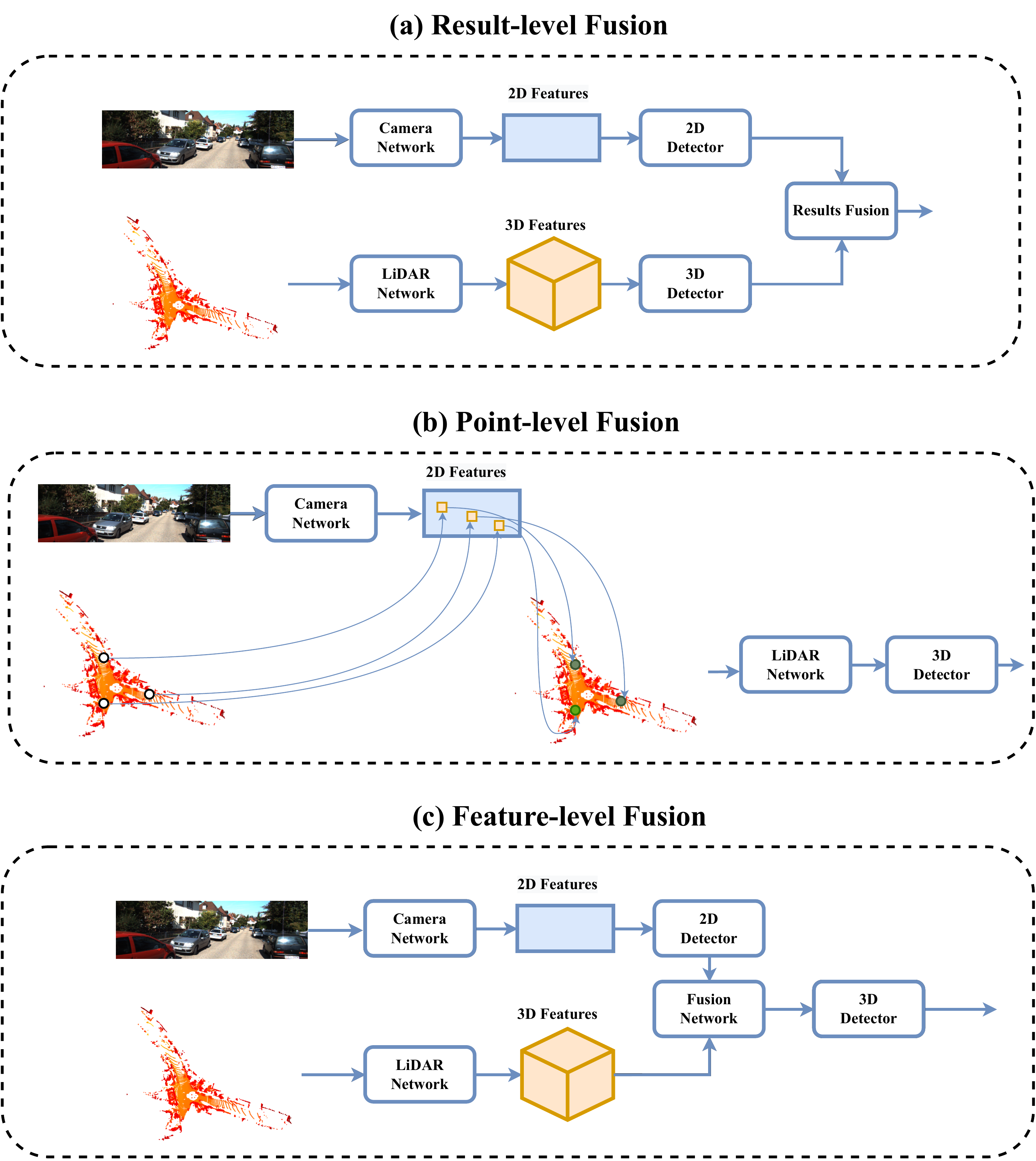}
      \caption{Multi-sensor fusion methods are divided into three types: result-level, point-level and feature-level fusion. (\textbf{a}) Result-level fusion fuses the output of individual detectors. (\textbf{b}) Point-level fusion projects the point clouds onto the image and acquires the corresponding features for fusion. (\textbf{c}) Feature-level fusion starts by acquiring features from different modalities and then fusing the features.}
      \label{figurelabel1}
\end{figure}

With the development of autonomous driving, 3D object detection is getting more and more attention. Its goal is to classify and locate objects in 3D space accurately. Recently, LiDAR and cameras have been the primary sources of information for 3D object detection.

As point clouds can provide accurate depth information about objects, in the beginning, researchers mainly used LiDAR-based methods \cite{VoxelNet, SECOND, PointPillars} for object detection. However, point clouds have less color or semantic information due to sparsity. As a result, small, distant, or obscured objects are challenging to detect in the single LiDAR mode because of the lack of sufficient points.

Images, on the other hand, can be a powerful addition to LiDAR information because they have a lot of semantic information but no depth information. Therefore, multi-modal fusion is of great importance for accurate and reliable perception.

As shown in Fig \ref{figurelabel1}, existing LiDAR-camera fusion methods can be broadly classified into result-level, feature-level, and point-level fusion. Result-level fusion methods \cite{FrustumCS, CLOCs} fuse the results of different detectors. The detection results of each sensor limit their detection effect, so they are rarely used nowadays. This paper focuses on illustrating the other two categories.

Point-level fusion methods \cite{EPNet, MVXNet, MultimodalVP, PointAug} establish links between point clouds and pixels, primarily through calibration matrices, so that data of one kind can be projected into another space, enhancing the data representation. Although some success has been achieved, the coarse mapping relationships inevitably introduce a degree of interference into the network. Also, due to the difference between point clouds' sparsity and pixels' density, only a little information will complete the matching fusion, resulting in a severe loss of information. Feature-level fusion methods\cite{MV3D, BEVFusion1, BEVFusion2, HomogeneousMF} use different modal networks to obtain the corresponding modal data features and subsequently fuse them. These methods have less information loss but make it hard to reconstruct the correspondence precisely between different modal features. 

To address the above issues, we propose a generic multi-modal 3D detection framework, i.e., MMFusion, in this paper. It contains two separate data streams: the LiDAR and Camera streams. To improve the LiDAR stream's performance, a plug-and-play module called the Voxel Local Perception Module (VLPM) has been designed so that voxels can fully express internal local information. Furthermore, the feature output from both streams is fused by the Multi-modal Feature Fusion Module (MFFM), and the output is then obtained using the detection head.

Specifically, the detection framework proposed here employs different data streams to encode the raw inputs into features within their respective feature spaces. As our framework is a generic approach, it can incorporate the current single-modal models of LiDAR and cameras into its respective streams. We also find that the voxel features previously obtained using local averages \cite{VoxelNet, SECOND, PointPillars} do not accurately reflect the actual voxel features. This motivates us to design a dynamic selection module for local features based on the attention mechanism, which can autonomously select the features of interest to enhance the representation of the model. In order to find precise connections between point cloud features and image features, we introduce MFFM to fuse features of different modalities. It avoids the direct use of hard or coarse fusion using geometric relations in previous approaches, and adaptively selects the desired part for fusion. Extensive experiments on the KITTI open dataset have shown that our approach can perform better than most current methods. Our contributions are summarized as follows:
\begin{itemize}
\item	A generic detection network framework for multi-modal data fusion (i.e., LiDAR-camera fusion) is proposed that can leverage multiple existing network architectures.
\item	A plug-and-play module VLPM is proposed for fully acquiring local features during voxelization, which enhances the representability of voxel grids.
\item	We propose a multi-modal fusion module MFFM to adaptively select different parts of different modal features for fine-grained fusion, which can make full use of the complementary properties of multi-modal features.
\item	Our framework MMFusion outperforms existing benchmarks and improves their detection on the KITTI dataset, especially for small objects.
\end{itemize}

\section{Related Works}

\subsection{LiDAR-based 3D Object Detection}

\textbf{Voxel-based methods.} Voxel-based methods first voxelize the original point clouds using local averages \cite{VoxelNet, SECOND, PointPillars}, then utilize 3D convolution kernel \cite{SECOND} to extract voxel features into the bird's eye view (BEV), and finally use a one-stage detection head \cite{SECOND, Centerbased, CVF, SA-SSD} to detect objects on projected BEV plane. Some other works add a refinement network \cite{PVRCNN, PVRCNN++, VoxelR-CNN, LiDARR-CNN, EPNet, EPNet++, Density-Aware} to the existing one-stage detection network. Recent works \cite{PointTrans, ct3d, VISTA} employ the attention mechanism for feature extraction and target detection. These methods are faster but do not exploit all the original information because of voxelization.

\textbf{Point-based methods.} Point-based methods take the original point clouds directly as input, then apply the point cloud encoder \cite{PointNet++} for feature extraction to feed into the detection network \cite{PointRCNN, STD, NotAllPoints}. Point-based approaches are usually computationally expensive as the encoder directly processes the original point clouds.

\subsection{Image-based 3D Object Detection}

LiDAR sensors are expensive compared to cameras, which are easily available and inexpensive. In recent years, image-based 3D object detection has been considerably explored. Wang et al. \cite{GeometryUP, FCOS3D, AutoShape} extend image detectors with additional 3D branches using monocular images. With the advent of autonomous driving datasets \cite{nuScenes} and more camera sensors, works \cite{DETR3D, SIDE, LIGAStereo} with multi-view images are gradually emerging. Due to the lack of direct depth information in the images, researchers have found that accurately predicting object depth from object and scene cues becomes a key constraint on detection capabilities. The methods \cite{CADDN, ProgressiveCT} rely on a depth estimation network to extract implicit depth information and thus improve the network's effectiveness.

\begin{figure*}[thpb]
      \centering
      
      \includegraphics[width=0.9\textwidth]{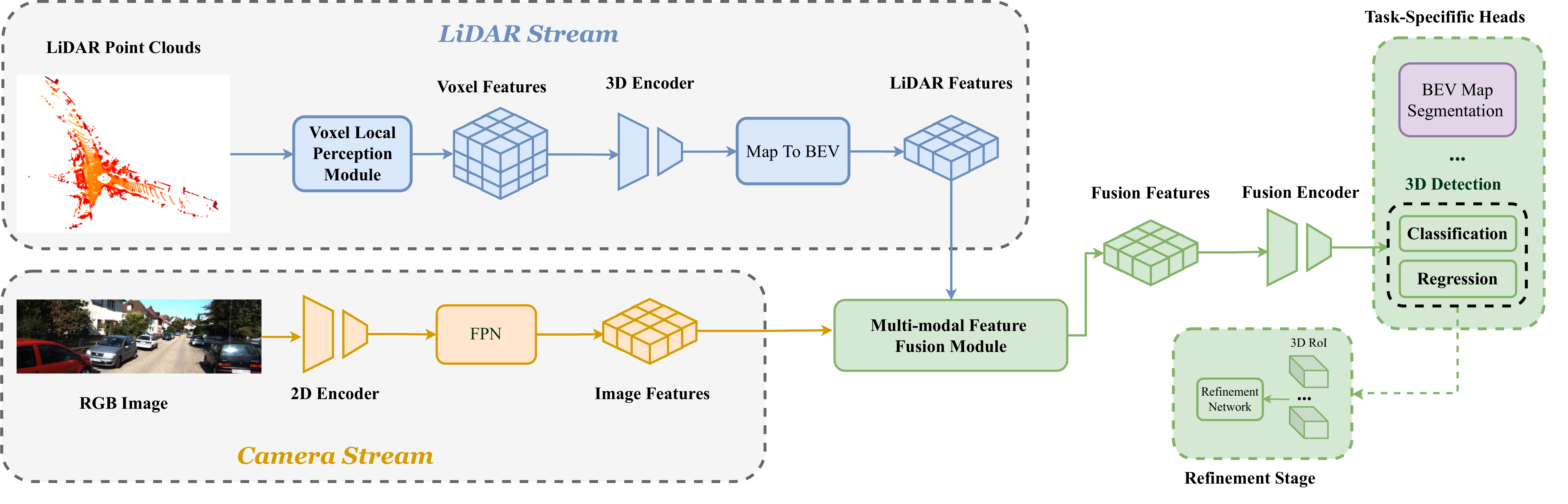}
      \caption{MMFusion consists of separate point clouds and image data streams, a multi-modal fusion module and a 3D object detection head. (\textbf{a}) LiDAR Stream extracts the original point cloud features.  (\textbf{b}) Image Stream extracts the RGB images features.  (\textbf{c}) Multi-modal Data Fusion Module fuses multi-modal data features from two streams.  (\textbf{d}) Fusion features support different tasks (e.g. 3D Object Detection) with task-specific heads.}
      \label{figurelabel2}
   \end{figure*}

\begin{figure*}[thpb]
      \centering
      
      \includegraphics[width=0.75\textwidth]{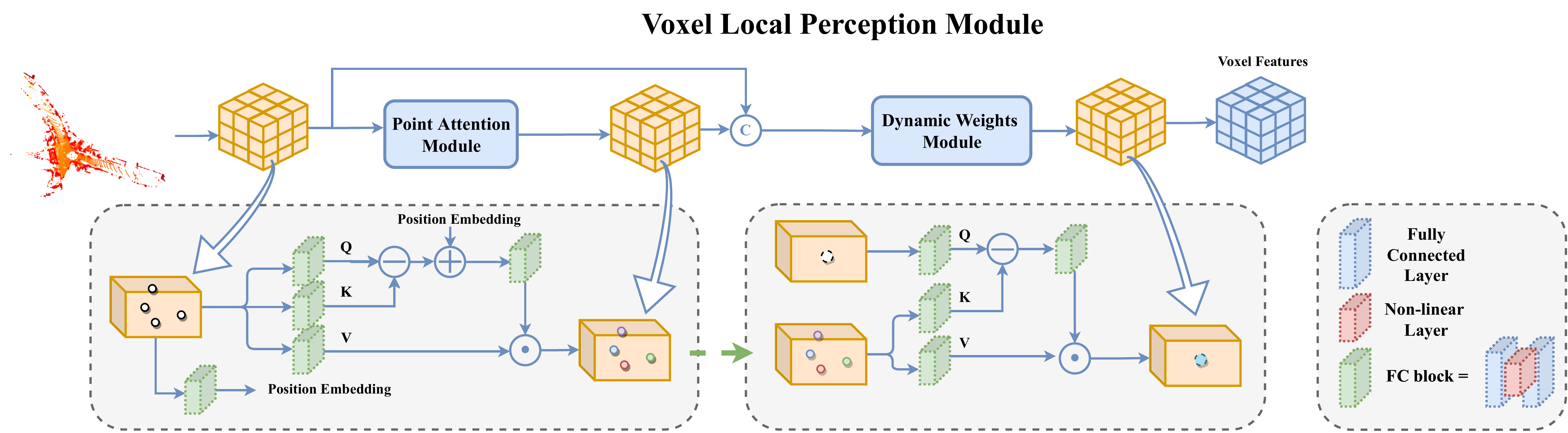}
      \caption{The structure of the Voxel Local Perception Module consists of Point Attention Module and Dynamic Weights Module. (\textbf{a}) Point Attention Module adaptively acquires the features of other points within the same voxel grid. (\textbf{b}) Dynamic Weights Module dynamically obtains the weights of different points of the same voxel to obtain the voxel characteristics.}
      \label{figurelabel3}
   \end{figure*}

\subsection{Multi-modal 3D Object Detection}

Due to the complementary nature of the information in point clouds and images, multi-sensor fusion methods are gradually gaining attention. This paper focuses on point-level and feature-level fusion shown in Fig \ref{figurelabel1} due to limitations in the performance of individual detectors.

\textbf{Point-level fusion methods.} Methods \cite{EPNet, MVXNet, MultimodalVP, PointAug} typically combine LiDAR points with image features at the corresponding location and apply LiDAR-based detection to the decorated point clouds. This fusion relies on coarse positional correspondence and has a high information loss \cite{BEVFusion1} due to the sparsity of the point clouds.

\textbf{Feature-level fusion methods.} These methods first exploit different networks to obtain features of different modal data and then use the fusion network to obtain fused features for detection. Specifically, Chen et al. \cite{MV3D, AVOD} utilize coarse geometric relationships to fuse features, obtaining suboptimal fusion features. Liu et al. \cite{BEVFusion1, BEVFusion2, HomogeneousMF} exploit a transformation network to construct 3D features of multi-view images before fusing them with point-level element features. However, the switching network has additional computing costs due to the redundant spatial information in the point cloud itself. Furthermore, an attention mechanism is employed in \cite{DeepFusion, TransFusion} to get different weights for features to be fused.

\section{Method}

In this section, we first detail the overall structure of our network in Sec.~\ref{sectionA}. Then, the Voxel Local Perception Module and Multi-modal Feature Fusion Module will be elaborated in Sec.~\ref{sectionB} and Sec.~\ref{sectionD}. The other parts of the network will be explained in the remaining subsections.

\subsection{Overall Structure}
\label{sectionA}
As shown in Fig \ref{figurelabel2}, the network mainly consists of independent point clouds and image data streams, a multi-modal fusion module, and a 3D object detection head. Given data input from different modalities, we design specific encoders to extract the respective features efficiently. Then we use the Multi-modal Data Fusion Module to dynamically extract different modal features, which can alleviate the misalignment problem of different modal features. Rather than using deep building networks, the network can achieve faster detection speeds. More details are elaborated in the following subsections.

\subsection{Voxel Local Perception Module}
\label{sectionB}
Due to the sparse characteristic of point clouds, extracting features directly from point clouds is computationally intensive. Therefore, methods \cite{VoxelNet, SECOND, PointPillars} first voxelize point clouds and then extract features from them. The details are as follows:

The point clouds in the range $(W, H, D)$ are first divided into voxel meshes of the same size, each with a size of $(w, h, d)$. To facilitate processing, set the number of point clouds sampled in each voxel grid as $N$. Finally, a maximum pooling operation or an average pooling operation is performed for each voxel grid.

\begin{figure*}[thpb]
      \centering
      \includegraphics[width=0.85\textwidth]{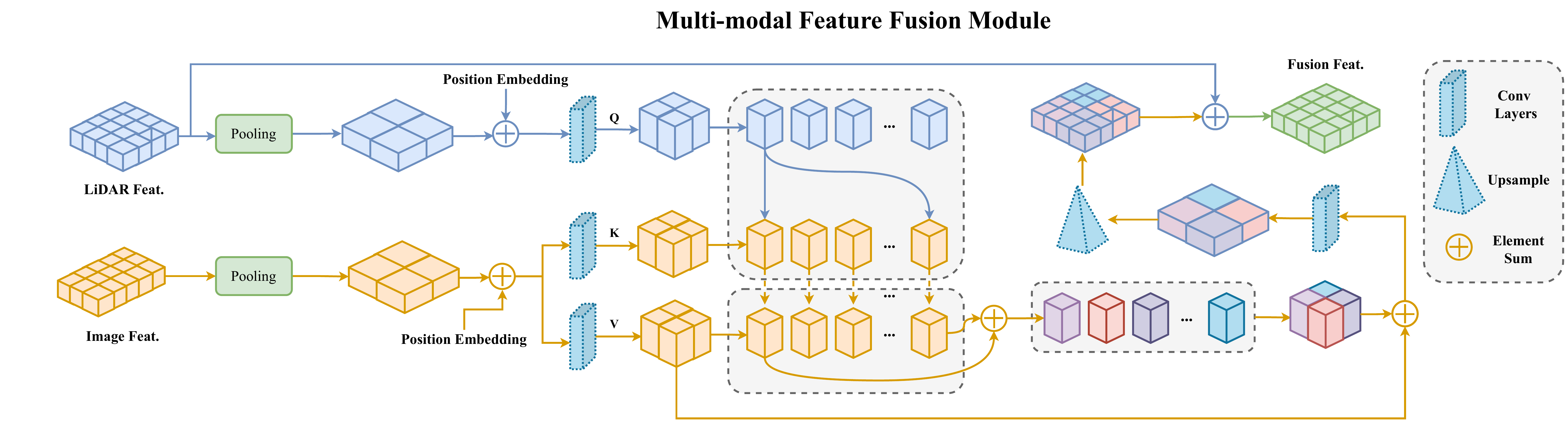}
      \caption{The structure of Multi-modal Feature Fusion Module. (\textbf{a}) It first compresses different modal features to a uniform size. (\textbf{b}) Then the features are mapped to the same feature space. (\textbf{c}) Finally, it uses an element-wise sum to fuse features of the same feature space.}
      \label{figurelabel4}
\end{figure*}

Although effective, direct pooling of voxel grids inevitably results in a significant loss of local information. Using only one pooling operation is not enough to accurately obtain the local features within a voxel grid, and the weight of each point within the voxel grid is also an important aspect to think about. As a result, in order to retain as much of the required point information as possible, we propose the Voxel Local Perception Module (VLPM), as shown in Fig \ref{figurelabel3}. Our Voxel Local Perception Module consists of Point Attention Module and Dynamic Weights Module.

\textbf{Point Attention Module.} To adaptively obtain local information from other points in the same voxel grid, we design a simple but effective module inspired by \cite{PointTrans}. 

As shown in Fig \ref{figurelabel3}, an FC block contains two linear layers and a ReLU non-linear layer. For each voxel grid $\mathcal{V}_k\in \mathbb{R}^{N \times 4}$, we use FC blocks $\alpha$ and $\beta$ that act on the spatial coordinates $c_i=[x_i, y_i, z_i]$ of each point to obtain Query $Q_i$ and Key $K_i$ of the self-attention mechanism. At the same time, point features $v_i\in V_k$ are fed into another FC block $\gamma$ to obtain Value $V_i$, which is
$$
Q_i = \alpha(c_i),  K_i = \beta(c_i), V_i = \gamma(p_i). \eqno{(1)}
$$
Afterward, we use an FC block $\epsilon$ to establish the similarity link $w_{ij}$ between the Query $Q_i$ and Key $K_j$ of points $p_i$ and $p_j$ under the same voxel grid (i.e., using additive attention to obtain the correlation coefficient):
$$
w_{ij} = \epsilon(Q_i - K_j + p_{ij}), \eqno{(2)}
$$
where $p_{ij}$ represents the position encoding between $p_i$ and $p_j$ from an FC block $\delta$:
$$
p_{ij} = \delta(c_i - c_j). \eqno{(3)}
$$
Consequently, for points $p_j$ in the same voxel grid $\mathcal{V}_k$, one can get the new features $f_i$ of point $p_i$:
$$
f_i = \sum_{p_j \in \mathcal{V}_k} w_{ij} \odot V_j. \eqno{(4)}
$$
\textbf{Dynamic Weights Module.} Different points within the same pixel grid usually have different weights (e.g., object edge points require a higher weight than background points). Inspired by this observation, we dynamically obtain the corresponding weights based on the point's coordinate values, as shown in Fig \ref{figurelabel3}. 

Unlike Point Attention Module, we get the weight $w_{i}$ in the current voxel grid $\mathcal{V}_k$ based on the position $c_i$ and $\bar{c}$ of each point $p_i$ and the average point $\bar{p}=[\bar{x}, \bar{y}, \bar{z}, ...]$:
$$
w_i = \theta(\zeta(\bar{c}) - \eta(c_i)). \eqno{(5)}
$$
where $\theta$, $\zeta$, and $\eta$ represents the FC blocks.

Then we obtain the features $f_{\mathcal{V}_k}$ of the current voxel grid $\mathcal{V}_k$. That is, 
$$
{f_{\mathcal{V}_k}} = \sum_{p_i \in \mathcal{V}_k} w_i \odot p_i.  \eqno{(6)}
$$

Here, the Dynamic Weights Module does not utilize the position coding information since, generally speaking, the Point Attention Module has already given the point clouds location information.

\subsection{Data Streams}
Different data stream architectures extract data features from different modalities separately.

\textbf{LiDAR Stream.} LiDAR point clouds can provide accurate spatial information. Objects on the road do not generally overlap vertically, so the projection of features onto the BEV plane does not cause interference. The methods \cite{SECOND, PointPillars} of extracting voxel features to obtain BEV features have become the de-facto standard.

Our framework can incorporate any feature extraction network to map the 3D features onto a BEV 2D space. We follow the design of the most widely used architectural networks \cite{SECOND, PVRCNN, PVRCNN++} to build our backbone networks.

\textbf{Camera Stream.} Camera images can provide high-resolution information on shapes and textures. For example, distant and small objects are difficult to detect due to the sparseness of the point clouds, but they are still clear and distinguishable in images.

Like the LiDAR stream, any 2D feature extraction network can be added to our framework to extract images' rich semantic and textural information. Instead of the widely used convolutional neural network ResNet \cite{Resnet}, we use an attention-based representative network called Swin Transformer \cite{SwinTrans} as the backbone of the camera stream, demonstrating excellent performance in visual tasks and significantly setting previous records. Following the practice \cite{FPN}, we then use a standard Feature Pyramid Network (FPN) to obtain multi-scale features.

\begin{table*}[h]
\caption{Comparison with state-of-the-art methods on the KITTI
test set for car 3D detection. The results are evaluated with AP$|_{R_{40}}$. All results are collected from the KITTI official website.}
\label{table_1}
\begin{center}
\renewcommand{\arraystretch}{1.2}
\begin{tabular}{cccccccccccc}
\hline

\hline
\multirow{2}{*}{Method}
&\multirow{2}{*}{Stage}
& \multicolumn{3}{c}{Car} 
& \multicolumn{3}{c}{Cyclist}    
& \multicolumn{3}{c}{Pedestrian} 
&\multirow{2}{*}{3DmAP}
\\
\cline{3-11}
&&Easy&Mod.&Hard&Easy&Mod.&Hard&Easy&Mod.&Hard\\
\hline
\multicolumn{12}{c}{LiDAR} \\
\hline
SECOND\cite{SECOND} &One&85.29 &76.60 &71.77  &71.05 &55.64 &49.83 &43.04 &35.92 &33.56&58.08\\
PointPillars\cite{PointPillars} &One&82.58&74.31&68.99   &77.10&58.65&51.92  &51.45&41.92&38.89&60.65\\
PointRCNN\cite{PointRCNN} &Two&86.96 &75.64 &70.70 &74.96 &58.82 &52.53 &47.98 &39.37 &36.01&60.33 \\
STD\cite{STD} &Two&87.95 &79.71 &75.09 &78.69 &61.59 &55.30 &53.29 &42.47 & 38.35&63.60 \\
Part $A^2$\cite{parta2}&Two &87.81 &78.49 &73.51 &79.17 &63.52 &56.93 &\textbf{53.10} &43.35 & 40.06 &63.99\\
3DSSD\cite{3dssd} &One&88.36 &79.57 &74.55 &\textbf{82.48} &64.10 &56.90 &50.64 &43.09 & 39.65 &64.37\\
SA-SSD\cite{SA-SSD} &One&88.75 &79.79 &74.16 &- &- &- &- &- & - &-\\
PV-RCNN\cite{PVRCNN}&Two &90.25 &81.43 &76.82 &78.60  &63.71 &57.65 &52.17 &43.29 &40.29 &64.91\\
CT3D\cite{ct3d} &Two&87.83 &81.77 &77.16 &- &- &- &- &- & - &-\\
Voxel R-CNN\cite{VoxelR-CNN} &Two&90.90&81.62&77.06  &- &- &- &- &- & - &-\\
\hline
\multicolumn{12}{c}{LiDAR+RGB} \\
\hline
MV3D\cite{MV3D}&One&74.97&63.63&54.00 &- &- &- &- &- & - &-\\
CVFNet\cite{CVF}&One&88.75&77.70&71.95 &- &- &- &- &- & -&- \\
3D-CVF\cite{3DCVF}&One&89.20&80.05&73.11 &- &- &- &- &- & -&- \\
EPNet\cite{EPNet}&Two&89.81&79.28&74.59 &- &- &- &- &- & -&- \\

CLOCs\cite{CLOCs}&Two&89.16 &82.28 &77.23 &- &- &- &- &- & - &-\\
DVF\cite{DVF}&One&89.40 &82.45 &77.56 &- &- &- &- &- & -& -\\
EPNet++\cite{EPNet++}&Two&\textbf{91.37}\textsuperscript{\dag}&81.96&76.71&76.15&	59.71&53.67&52.79&44.38&\textbf{41.29} & 64.23\\
\hline
\rowcolor{green!10}Ours &Two& 89.05 & \textbf{82.50} & \textbf{77.59} & 80.82 & \textbf{65.39} & \textbf{59.10} & 51.39& 43.30 & 39.67 &\textbf{65.42}\\
\hline

\hline
\end{tabular}

\end{center}
\noindent{\footnotesize{\textsuperscript{\ddag} The best results are in bold.}}
\end{table*}

\subsection{Multi-modal Feature Fusion Module}
\label{sectionD}
Different sensors acquire data from different views; therefore, different modal features are in different feature spaces. For instance, LiDAR stream features are in BEV space, and camera stream features are in RV space (i.e., front, back, left, right).

In this case, the same element in different modal features might correspond to entirely different spatial locations. So we cannot use element-level operations (like element-wise sum) to directly combine feature maps from different modalities. As such, a Multi-modal Feature Fusion Module, which uses the attention mechanism, is proposed here to transform different features into a unified feature space flexibly.

\textbf{Overall Architecture.} 
As shown in Fig~\ref{figurelabel4}, we make the LiDAR features dynamically fuse different regional features of interest in the image so that the LiDAR and image features are in the same feature space. Specifically, we use the attention mechanism to construct a correlation of regional features across modalities.

After the pooling and adding the position encoding $p_{Le}$ and $p_{Ie}$, LiDAR features $f_L\in\mathbb{R}^{C \times H_1 \times W_1}$ and image features $f_I\in \mathbb{R}^{C \times H_2 \times W_2}$ first go through the pooling layer $P$ to obtain features $f_{L^{'}}$ and $f_{I^{'}}$ of the same size: 
$$
f_{L^{'}} = P(f_{L})+p_{Le},   f_{I^{'}} = P(f_{I})+p_{Ie}. \eqno{(7)} 
$$

Moreover, they will become the Query $Q\in \mathbb{R}^{n \times C}$, Key $K\in\mathbb{R}^{m \times C}$, and Value $V\in\mathbb{R}^{m \times C}$ by passing convolution layers $\phi$, $\psi$, and $\vartheta$ as below.
$$
Q = \phi(f_{L^{'}}), K = \psi(f_{I^{'}}), V =\vartheta(f_{I^{'}}). \eqno{(8)} 
$$

Because of the large size of the feature map, we use scaled dot-product attention presented in \cite{transformer} instead of additive attention to calculate the correlation coefficient matrix $W\in [0, 1]^{n \times m}$ and speed up the computation, as:
$$
W = softmax(\frac{QK^{\mathrm{T}}}{\sqrt{C}}).   \eqno{(9)} 
$$
where $C$ is the dimensional size of $K$.

At this step, we can obtain the corresponding image features $f_{I^{''}}$ of LiDAR features by a convolution layer $\gamma$ and fuse them with an upsampling layer $\epsilon$ and simple addition for fusion features $f_{F}$, i.e.,
$$
f_{F} = f_{L} + \epsilon(f_{I^{''}}),  f_{I^{''}} = \gamma(WV + V). \eqno{(10)} 
$$

\textbf{Pooling Operation.} 
Usually, the product of $H_1W_1$ and $H_2W_2$ is large, causing the network's training to be prolonged. It is also found that too many feature sequences are not conducive to extracting key features, as shown in Sec~\ref{Ablation}. 

Hence, we first pool the features $f_L, f_I$ to smaller sizes $(H_1^{'}, W_1^{'}), (H_2^{'}, W_2^{'})$. The fused features $f_F$ are then scaled to their original size $(H_1, W_1)$ and then added to LiDAR features $f_L$.

\subsection{Detection Head and Loss Function}
After the series of operations mentioned above, we can obtain the fused feature map. The network generates results through the downstream task-specific heads. In the field of 3D target detection, we use the detection head of \cite{SECOND, Density-Aware, VoxelR-CNN} to verify the validity of our network.

\begin{table*}[h]
\caption{3D object detection results on the KITTI validation set for SECOND, PDV and Voxel-RCNN as baseline. The results are evaluated with AP$|_{R_{40}}$.}
\label{table_2}
\begin{center}
\renewcommand{\arraystretch}{1.2}
\begin{tabular}{ccccccccccc}
\hline

\hline
\multirow{2}{*}{Method}
& \multicolumn{3}{c}{Car}
& \multicolumn{3}{c}{Cyclist}    
& \multicolumn{3}{c}{Pedestrian} 
&\multirow{2}{*}{3DmAP}
\\
\cline{2-10}
&Easy&Mod.&Hard&Easy&Mod.&Hard&Easy&Mod.&Hard\\
\hline
SECOND\textsuperscript{\dag}\cite{SECOND}  & 90.11 & 81.08 & 78.11
& 85.53 & 68.58 & 64.45 & 57.67 & 51.92 & 47.02 & 69.38\\
MMFusion+SECOND\textsuperscript{$\star$} & \textbf{91.01}\textsuperscript{\ddag} & \textbf{81.51} & \textbf{78.68} & \textbf{87.53} & \textbf{69.72} & \textbf{65.32} & \textbf{59.29} & \textbf{54.52} & \textbf{49.67} & \textbf{70.8}\\ 
\rowcolor{green!10}Improvement& +\textit{0.9} & +\textit{0.43} & +\textit{0.57}
& +\textit{2.00} & +\textit{1.14} & +\textit{0.87} & +\textit{1.62} & +\textit{2.60} & +\textit{2.65} & +\textit{1.42}\\
\hline
PDV\textsuperscript{\dag}\cite{Density-Aware} & 92.23 & 85.04 & 82.66 & 89.9 & 71.73 & 68.5
& 65.14 & 57.64 & 52.66 & 73.94\\
MMFusion+PDV & 91.48 & 84.13 & 82.2 & \textbf{91.86} & \textbf{74.39} & \textbf{70.07} & 64.87 & \textbf{58.45} & \textbf{53.89} & \textbf{74.59} \\
\rowcolor{green!10}Improvement& -\textit{0.75} & -\textit{0.91} & -\textit{0.46}
& +\textit{1.96} & +\textit{2.66} & +\textit{1.57} & -\textit{0.27} & +\textit{0.81} & +\textit{1.23} & +\textit{0.65}\\
\hline
Voxel R-CNN\textsuperscript{\dag}\cite{VoxelR-CNN} & 92.26 & 84.98 & 82.51 & 92.24 & 73.66 & 70.21 & 64.86 & 58.65 & 53.98 & 74.82 \\ 
MMFusion+Voxel R-CNN & 92.21 & 83.14 & 82.4 & \textbf{92.59} & \textbf{74.39} & \textbf{70.79} & \textbf{66.67} & \textbf{59.78} & \textbf{54.57} & \textbf{75.17}\\
\rowcolor{green!10}Improvement& -\textit{0.05} & -\textit{1.84} & -\textit{0.11}
& +\textit{0.35} & +\textit{0.73} & +\textit{0.58} & +\textit{1.81} & +\textit{1.13} & +\textit{0.59} & +\textit{0.35}\\
\hline

\hline
\end{tabular}
\end{center}
\noindent{\footnotesize{\textsuperscript{\dag} The detection results of the benchmark are reproduced by its  official released code.}}

\noindent{\footnotesize{\textsuperscript{\ddag} The improved results are in bold.}}

\noindent{\footnotesize{\textsuperscript{$\star$} Use the current network's detection head.}}
\end{table*}

\begin{table}[h]
\setlength{\abovecaptionskip}{0pt}
\setlength{\belowcaptionskip}{10pt}
\caption{Effects of proposed components on the KITTI validation set.}
\label{table_3}
\begin{center}
\scalebox{0.8}{
\renewcommand{\arraystretch}{1.2}
\begin{tabular}{ccccccc}
\hline

\hline
\multirow{2}{*}{Baseline}
& \multirow{2}{*}{VLPM}
& \multirow{2}{*}{MFFM}
& \multirow{2}{*}{Two-stage}
& \multicolumn{3}{c}{3D}    
\\
\cline{5-7}
&&&&Car&Cyclist&Pedestrian\\
\hline
\checkmark	&&&	& 83.1 & 72.85 & 52.20\\
\checkmark&\checkmark&&	& 83.78 & 71.83 & 54.01\\
\checkmark&&\checkmark&	& 84.22 & 73.40 & 54.36\\
\checkmark&\checkmark&\checkmark&	& \textbf{83.73}\cellcolor{green!10}(+\textit{0.63}) & \textbf{74.19}\cellcolor{green!10}(+\textit{1.34}) & \textbf{54.49}\cellcolor{green!10}(+\textit{2.29})\\
\checkmark&\checkmark&\checkmark&\checkmark& \textbf{85.92}\cellcolor{green!10}(+\textit{2.82}) & \textbf{79.26}\cellcolor{green!10}(+\textit{6.41}) & \textbf{60.34}\cellcolor{green!10}(+\textit{8.14})\\
\hline

\hline
\end{tabular}
}
\end{center}
\end{table}

For the one-stage detection head, following common standard \cite{SECOND}, we adopt the Region Proposal Network (RPN) loss $\mathcal{L}_{RPN}$, which consists of anchor classification loss $\mathcal{L}_{cls}$, regression loss $\mathcal{L}_{reg}$, and direction classification loss $\mathcal{L}_{dir}$. In short,
$$
\mathcal{L}_{total} = \mathcal{L}_{RPN} = \mathcal{L}_{cls} + \alpha\mathcal{L}_{reg} + \beta\mathcal{L}_{dir},
\eqno{(11)} 
$$
where $\alpha$ is set to 2.0 and $\beta$ is set to 0.2.

For the two-stage detection head, Proposal Refinement Network (PRN) loss $\mathcal{L}_{PRN}$ will be part of the loss function following \cite{VoxelR-CNN}. In brief,
$$
\mathcal{L}_{total} = \mathcal{L}_{RPN} + \mathcal{L}_{PRN}. \eqno{(12)} 
$$

\section{Experiments}
\subsection{Dataset}

KITTI \cite{KITTI} has now been widely used as a benchmark for autonomous driving datasets. It contains three categories in total: cars, cyclists, and pedestrians. Each category is divided into three difficulty levels: easy, medium, and hard, based on the size, occlusion level, and truncation level of the objects. The total dataset contains 7481 training samples and 7518 test samples. For validation purposes, the training samples are typically divided into a training set of 3712 and a validation set of 3769. We use the training set for the experimental study of the validation test and test set.

\subsection{Setup Details}

\textbf{Data augmentation.} 
Data augmentation is performed to improve the generalization performance of the network. Following \cite{VoxelR-CNN}, we first extract some ground truth boxes and place them at random locations. Then, the point clouds are randomly flipped along the $x$-axis, rotated along the $z$-axis in the range [-$\pi$/4, $\pi$/4], and scaled in the range [0.95, 1.05].

\textbf{Input Parameters.}
Since KITTI only provides annotations in the front camera's field of view. We set the range of point clouds to [0, 70.4m], [-40m, 40m], [-3m, 1m] in the $(x, y, z)$ axis following \cite{SECOND}. The input voxel size is set to (0.05m, 0.05m, 0.1m). So the size of the whole 3D space after voxelization is 1600 × 1408 × 40. The maximum number of non-empty voxels is
set to 16,000 in training and 40,000 in testing. We reshape the dimensions of the images to 3 × 1216 × 352 and take them as the input to the image stream.

\textbf{Network Architecture.} The backbone of LiDAR Stream is based on SECOND \cite{SECOND}. We add the Voxel Local Perception Module with two stages before the backbone network to acquire voxel features. The 3D backbone has four stages with filter numbers of 16, 32, 48, and 64. We then utilize two 2D network blocks to process the features. The first block has the same X and Y resolution as the 3D backbone output, while the second block has half the resolution of the first. Moreover, the LiDAR Stream features have a size of 256 × 200 × 176.

\begin{table}[h]
\setlength{\abovecaptionskip}{0cm}
\setlength{\belowcaptionskip}{10pt}
\caption{Effects of different pooling sizes on the KITTI validation set.}
\label{table_4}
\begin{center}
\renewcommand{\arraystretch}{1.2}
\begin{tabular}{ccccc}
\hline

\hline
Size & Car&Cyclist &Pedestrian & 3DmAP\\
\hline
\rowcolor{green!10}(25, 22)	&\textbf{83.73} & \textbf{74.19} & \textbf{54.49} & \textbf{70.8}\\
(50, 44)	&83.39 & 73.59 & 54.67 & 70.55\\
(100, 88)	&83.46 & 71.69 & 52.9 & 69.35\\
\hline

\hline
\end{tabular}
\end{center}
\end{table}

\begin{figure*}[thpb]
      \centering
      \includegraphics[width=0.8\textwidth]{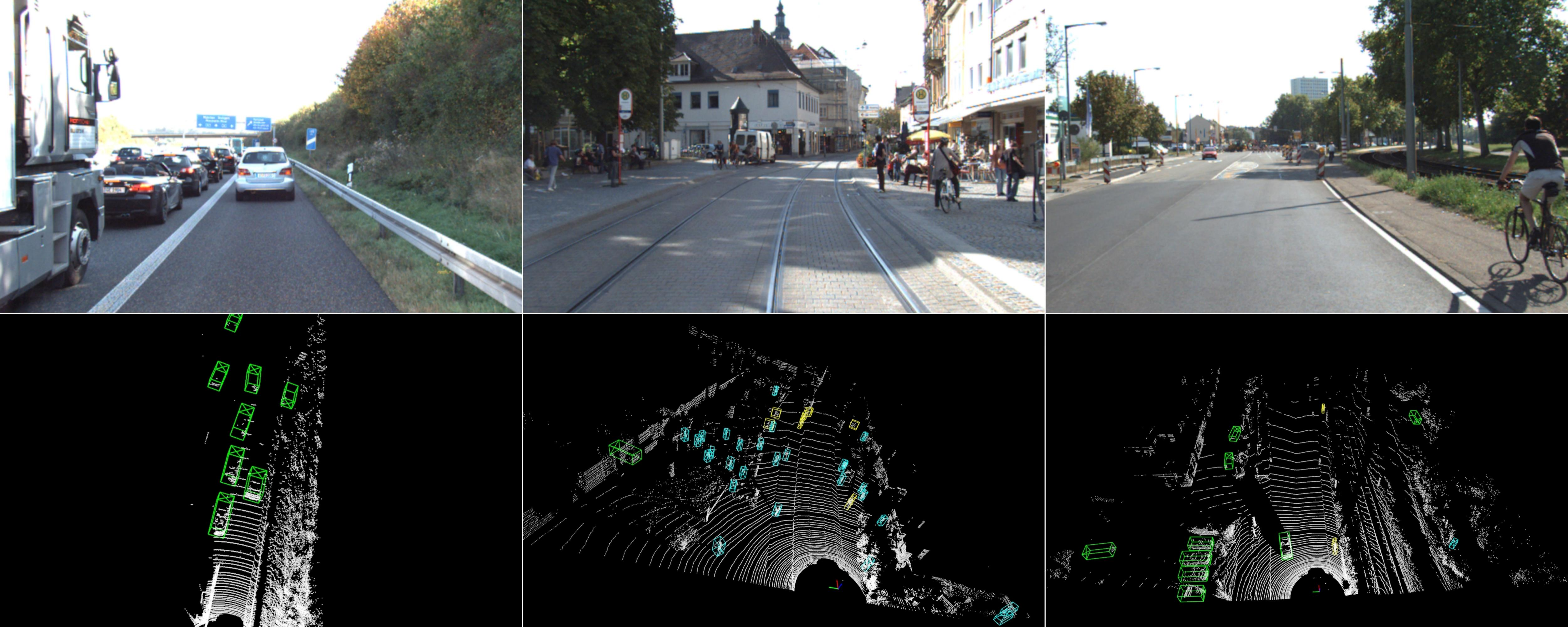}
      \caption{The visualization and detection results on the KITTI dataset. We use green, yellow and blue 3D bounding boxes to represent the detection results of cars, cyclists and pedestrians.}
      \label{figurelabel5}
\end{figure*}

We employ Swin Transformer \cite{SwinTrans} with four blocks as the backbone of the Camera Stream. The window size is changed to $M=5$, and the channel number of the first stage's hidden layers is changed to $C=32$. The feature dimension of each block of output is 32, 64, 128, and 256, and the size becomes one-half of its original size with each passing block. After that, we leverage a Feature Pyramid Network \cite{FeaturePN} to assemble them and grab the image features with a size of 256 × 39 × 11.

Data from different modalities is fed into the MFFM for fusion. Finally, downsampling and upsampling 2D networks are employed with feature dimensions (128, 256) to obtain fusion features whose size is 256 × 200 × 176. 

To verify the validity and generality of our framework, different detection heads \cite{SECOND, VoxelR-CNN, Density-Aware} are used in the experiments.

\textbf{Training.}
OpenPCDet \cite{openpcdet} is adopted as our codebase. All the models are trained on two 3090Ti GPU cards with a batch size of 4 and 80 epochs. We keep the same training settings as for the baseline detector. Specifically, the Adam optimizer is applied with a learning rate of 0.003 for \cite{SECOND} and 0.01 for \cite{VoxelR-CNN, Density-Aware}. The weight decay parameter of 0.01 is the same for all models.

\subsection{Comparison with the State of the Art}
We evaluate our detection network on the KITTI test set and then submit the results to the official online benchmark server. As shown in Table \ref{table_1}, 
our MMFusion outperforms PointRCNN \cite{PointRCNN}, PV-RCNN \cite{PVRCNN}, and EPNet++ \cite{EPNet++}  with 5.09$\%$, 0.51$\%$, and 1.19$\%$ AP$|_{R_{40}}$ respectively on the mean average precision (mAP). As a two-stage detection network, it not only shows good performance in detecting cars, but also performs well in small target detection, which usually needs improvement in LiDAR measurements. In particular, benefiting from the complementary multi-modal information, our network has significantly better detection of cyclists than other competitors.

In addition, we show the detection results on the KITTI validation set in Table \ref{table_2}. On the mAP, our MMFusion outperforms the baseline SECOND \cite{SECOND}, PDV \cite{Density-Aware}, and Voxel-RCNN \cite{VoxelR-CNN} by 1.42$\%$, 0.35$\%$, and 0.65$\%$ AP$|_{R_{40}}$, respectively. Notably, our MMFusion improves the cyclists' detection performance by 1.34$\%$, 0.55$\%$ and 2.06$\%$, respectively, further illustrating the effectiveness of information fusion. Due to the interference caused by the redundancy of fused information, the model's detection effect on cars on PDV and Voxel-RCNN has slightly declined. Still, this cost is acceptable for improving minor target detection effects. Some detection results are shown in Fig \ref{figurelabel5}, where the network also maintains good detection results in complex environments.

\subsection{Ablation Studies} \label{Ablation}
\textbf{Components.} By setting SECOND \cite{SECOND} as the baseline model, we examine the impact of each component of MMFusion on the network performance. Table \ref{table_3} shows that both VLPM and MFFM have a remarkable promotion effect on the model.

\textbf{Pooling Sizes.} We set different pooling sizes of LiDAR features in the MFFM, and Table \ref{table_4} shows that the smaller pooling size helps to obtain a better performance of the model.

\section{Conclusions}
In this paper, we propose a generic 3D detection framework (MMFusion) using multi-modal features. The framework aims to achieve accurate fusion between LiDAR and images to improve 3D detection. We created a voxel local perception module to better preserve internal information during voxelization. Benefiting from the decoupling of data streams, our framework can take full advantage of existing feature extraction and detection networks. A Multi-modal Feature Fusion Module is subsequently designed to selectively combine feature output from different streams to achieve accurate fusion. Extensive experiments have demonstrated that our framework not only achieves superior performance but also improves the detection of existing benchmarks, especially bicycle and pedestrian detection on KITTI benchmarks, with strong robustness and generalization capabilities. We plan to use the framework to adapt to more types of data and downstream tasks in the future. Hopefully, our work will stimulate more research into multi-modal fusion for autonomous driving tasks.

\addtolength{\textheight}{-4cm}   

\bibliographystyle{IEEEtran}
\bibliography{cite}

\end{document}